# Hierarchical Mixtures-of-Experts for Exponential Family Regression Models with Generalized Linear Mean Functions: A Survey of Approximation and Consistency Results

**Wenxin Jiang and Martin A. Tanner**
*Department of Statistics, Northwestern University*
*Evanston, IL 60208, USA*

## Abstract

We investigate a class of hierarchical mixtures-of-experts (HME) models where exponential family regression models with generalized linear mean functions of the form $\psi(\alpha + \mathbf{x}^T\beta)$ are mixed. Here $\psi(\cdot)$ is the inverse link function. Suppose the true response $y$ follows an exponential family regression model with mean function belonging to a class of smooth functions of the form $\psi(h(\mathbf{x}))$ where $h(\cdot) \in W_2^\infty$ (a Sobolev class over $[0,1]^s$). It is shown that the HME probability density functions can approximate the true density, at a rate of $O(m^{-2/s})$ in $L_p$ norm, and at a rate of $O(m^{-4/s})$ in Kullback-Leibler divergence. These rates can be achieved within the family of HME structures with no more than $s$-layers, where $s$ is the dimension of the predictor $\mathbf{x}$. It is also shown that likelihood-based inference based on HME is consistent in recovering the truth, in the sense that as the sample size $n$ and the number of experts $m$ both increase, the mean square error of the predicted mean response goes to zero. Conditions for such results to hold are stated and discussed.

## 1 Introduction

Both the Mixtures-of-Experts (ME) model, introduced by Jacobs, Jordan, Nowlan and Hinton (1991), and the Hierarchical Mixtures-of-Experts (HME) model, introduced by Jordan and Jacobs (1994), have received considerable attention due to flexibility in modeling, appealing interpretation, and the availability of convenient computational algorithms. In contrast to the single-layer ME model, the HME model has a tree-structure and can summarize the data at multiple scales of resolution due to its use of nested predictor regions. By the way they are constructed, ME and HME models are natural tools for likelihood-based inference using the Expectation Maximization (EM) algorithm [Jordan and Jacobs (1994) and Jordan and Xu (1995)], as well as for Bayesian analysis based on data augmentation [Peng, Jacobs and Tanner (1996)]. An introduction and application of mixing experts for generalized linear models (GLMs) are presented in Jordan and Jacobs (1994) and Peng, Jacobs and Tanner (1996).

Both ME and HME have been empirically shown to be powerful and general frameworks for examining relationships among variables in a variety of settings [Cacciatore and Nowlan (1994), Meilă and Jordan (1995), Ghahramani and Hinton (1996), Tipping and Bishop (1997) and Jaakkola and Jordan (1998)]. Despite the fact that ME and HME have been incorporated into neural network textbooks [e.g., Bishop (1995) and Haykin (1994) which features an HME design on the cover], there has been very little formal statistical justification [see Zeevi, Meir and Maiorov (1998)] of the methodology. In this paper we consider the denseness and consistency of these models in the generalized linear model context. Before proceeding we present some notation regarding mixtures and hierarchical mixtures of generalized linear models and one-parameter exponential family regression models.

Generalized linear models, which are natural extensions of the usual linear model, are widely used in statistical practice [McCullagh and Nelder (1989)]. One-parameter exponential family regression models [see Bickel and Doksum (1977), page 67] with generalized linear mean functions (GLM1) are special examples of the generalized linear models, where the probability distribution is totally determined by the mean function. In the regression context, a GLM1 model proposes that the conditional expectation $\mu(\mathbf{x})$ of a real response variable $y$ is related to a vector of predictors $\mathbf{x} \in \Re^s$ via a generalized linear function $\mu(\mathbf{x}) = \psi(\alpha + \beta^T \mathbf{x})$, with $\alpha \in \Re$ and $\beta \in \Re^s$ being the regres-sion parameters and $\psi^{-1}(\cdot)$ being the link function. Examples include the log link where $\psi(\cdot) = \exp(\cdot)$, the logit link where $\psi(\cdot) = \exp(\cdot)/\{1 + \exp(\cdot)\}$, and the identity link which recovers the usual linear model. The inverse link function $\psi(\cdot)$ is used to map the entire real axis to a restricted region which contains the mean response. For example, when $y$ follows a Poisson distribution conditional on $\mathbf{x}$, a log link is often used so that the mean is non-negative. In general, the GLM1 probability density function of $y$ conditional on $\mathbf{x}$ is totally determined by the conditional mean function $\mu(\mathbf{x})$, having the form $p(y; \mathbf{x}) = \exp\{a(\mu) + b(y) + yc(\mu)\}$, where $\mu = \mu(\mathbf{x}) = \psi(\alpha + \boldsymbol{\beta}^T \mathbf{x})$, and $a(\cdot)$, $b(\cdot)$ and $c(\cdot)$ are some fixed functions. Such models include Poisson, binomial and exponential regression models, as well as the normal and gamma regression models with dispersion parameters regarded as known.

A mixtures-of-experts model assumes that the total output is a locally-weighted average of the output of several GLM1 experts. It is important to note that such a model differs from standard mixture models [e.g., Titterington, Smith and Makov (1985)] in that the weights depend on the predictor. A generic expert labeled by an index $J$, proposes that the response $y$, conditional on the predictor $\mathbf{x}$, follows a probability distribution with density $p_J(y; \mathbf{x}) = \pi(h_J(\mathbf{x}), y) = \exp\{a(\mu_J) + b(y) + yc(\mu_J)\}$, where $\mu_J = \psi(h_J(\mathbf{x}))$ and $h_J(\mathbf{x}) = \alpha_J + \boldsymbol{\beta}_J^T \mathbf{x}$. The total probability density of $y$, after combining several experts, has the form $p(y; \mathbf{x}) = \sum_J g_J(\mathbf{x}) p_J(y; \mathbf{x})$, where the local weight $g_J(\mathbf{x})$ depends on the predictor $\mathbf{x}$, and is often referred to as a gating function. The total mean response then becomes $\mu(\mathbf{x}) = \sum_J g_J(\mathbf{x}) \mu_J(\mathbf{x})$. A simple mixtures-of-experts model takes $J$ to be an integer. An HME model takes $J$ as an integer vector, with dimension equal to the number of layers in the expert network. An example of the HME model with two layers is given in Jordan and Jacobs (1994), as illustrated in Figure 1. Note that the HME is a graphical model with a probabilistic decision tree, where the weights of experts reflect a recursive stochastic decision process. In Figure 1, adapted from Jordan and Jacobs (1994), the expert label $J$ is a two-component vector with each component taking either value 1 or 2. The total mean response $\mu$ is recursively defined by $\mu = \sum_{i=1}^{2} g_i \mu_i$ and $\mu_i = \sum_{j=1}^{2} g_{j|i} \mu_{ij}$, where $g_i$ and $g_{j|i}$ are logistic-type local weights associated with the "gating networks" for the choice of experts or expert groups at each stage of the decision tree, conditional on the previous history of decisions. Note that the product $g_i g_{j|i}$ gives a weight $g_J(\mathbf{x}) = g_i g_{j|i}$ for the entire decision history $J = (i, j)$. At the top of the tree is the mean response $\mu$, which is dependent on the entire history of probabilistic decisions and also on the predictor $\mathbf{x}$.

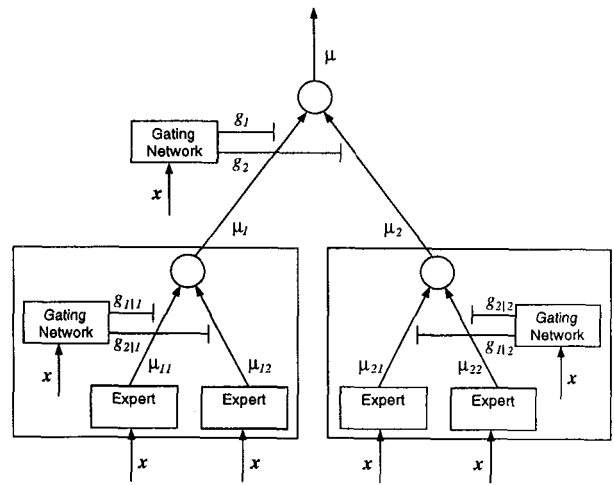

Figure 1: A Two-Layer Hierarchical Mixtures-of-Experts Model

One important issue is the approximation power of the HME models. Is the family of mean functions of the form $\sum_J g_J(\mathbf{x}) \mu_J(\mathbf{x})$ proposed by HME rich enough to approximate an arbitrary smooth mean function of a certain family to any degree of accuracy? What precision, in a certain norm, can the approximation based on a specific number of experts achieve? Such problems of denseness and complexity are well-described and studied in the neural network literature [see Mhaskar (1996)]. A different question is the consistent learning property of HME with respect to a specific learning procedure. An HME model, as later we will see, is characterized by a parameter vector, which can be estimated based on a training data set consisting of $n$ pairs of $(\mathbf{x}, y)$'s, following a learning procedure (or fitting method) such as the least-squares or the maximum likelihood approach. The consistency problem centers on whether the learning procedure will produce an estimated mean function being close to the true mean function when the size of the training data set is sufficiently large. Various methods of measuring the closeness include the convergence in probability and the convergence in mean square error of the estimated mean function. The latter is a stronger mode of convergence due to Chebyshev's inequality [see Bickel and Doksum (1977), page 463] and is the mode of convergence we will consider in this paper.

Regarding these important theoretical questions, it is demonstrated by Zeevi, Meir and Maiorov (1998) that one-layer mixtures of linear model experts can be used to approximate a class of smooth functions as the number of experts increases, and the least-squares method can be used to estimate the mean response consistently when the sample size increases. The goal of this paper is to extend this result to HME for GLM1s with nonlinear link functions, and to consider the consistency of maximum likelihood estimation. The maximum likelihood (ML) approach has two advantages over the conventional least square approach. (i) The maximum likelihood approach gives the smallest asymptotic variance for the estimator of the mean response, in the case of correct model specification. (ii) The convenient EM algorithm can be used naturally for maximizing the likelihood, just as in the case of ordinary mixture models. However there are two difficulties for studying the consistency properties of a likelihood-based approach. (i) The maximum likelihood method deals with density functions rather than with mean functions. A result on the denseness of mean functions, such as the one stated in Zeevi, Meir and Maiorov (1998), is not enough. We need to establish a similar result for the density functions. We show that HME for GLM1 density functions can be used to approximate density functions of the form $\pi(h(\mathbf{x}), y)$, where $h(\cdot)$ is an arbitrary smooth function. (ii) The maximum likelihood method minimizes the Kullback-Leibler (KL) divergence, while the consistency properties for the estimates of mean responses is usually investigated by showing that the mean square error (MSE) of the estimated mean responses converge to zero in some fashion. An extra condition is required to establish a relationship between the KL divergence of the *density functions* and the MSE, or the $L_2$ distance of the *mean functions*.

Finally, we note that the parameterization of the HME, as shown in the next section, is not identifiable. Care is needed for statements about the parameter estimates, which are not unique.

## 2 Notation and Definitions

### 2.1 The Family of Target Functions

Let $\Omega = [0,1]^s = \otimes_{q=1}^s [0,1]$, the space of the predictor $\mathbf{x}$, where $\otimes$ stands for the direct product. Let $A \subset \Re$ be the space of the response $y$. Let $(A, \mathcal{F}_A, \lambda)$ be a general measure space, $(\Omega, \mathcal{F}_\Omega, \kappa)$ be a probability space such that $\kappa$ is absolutely continuous with respect to the Lebesgue measure on $\Omega$, and $(\Omega \otimes A, \mathcal{F}_\Omega \otimes \mathcal{F}_A, \kappa \otimes \lambda)$ be the product measure space.

Consider the following target functions:

$$\varphi(\mathbf{x}, y) = \pi(h(\mathbf{x}), y) \qquad (1)$$

where $\pi(\cdot, \cdot) : \Re \otimes A \mapsto \Re$ is a fixed positive continuous function, $\pi(\cdot, y) : \Re \mapsto \Re$ is a continuously differentiable function for all $y$, and $\varphi : \Omega \otimes A \mapsto \Re$ is measurable $\mathcal{F}_\Omega \otimes \mathcal{F}_A$. In (1), $h : \Omega \mapsto \Re$ is assumed to have continuous second derivatives, $\sum_{\mathbf{k}: 0 \le |\mathbf{k}| \le 2} \|D^{\mathbf{k}} h\|_\infty \le$ 1, where $\mathbf{k} = (k_1, \ldots, k_s)$ is a $s$-dimensional vector of nonnegative integers between 0 and 2, $|\mathbf{k}| = \sum_{j=1}^s k_j$, $\|h\|_\infty \equiv \sup_{\mathbf{x} \in \Omega} |h(\mathbf{x})|$, and $D^{\mathbf{k}} h \equiv \frac{\partial^{|\mathbf{k}|} h}{\partial x_1^{k_1} \ldots \partial x_s^{k_s}}$. In other words, $h \in W_2^\infty$, where $W_2^\infty$ is a Sobolev space with sup-norm and second-order continuous differentiability.

We further assume that for all $h \in \Re$, $\pi(h, y)$ is a density function in $y$, satisfying $\int_A \pi(h, y) d\lambda(y) = 1$ and $\int_A y^2 \pi(h, y) d\lambda(y) < \infty$. A conditional mean function $\mu(\cdot)$, corresponding to a $\varphi(\cdot, \cdot)$, is defined by

$$\mu(\mathbf{x}) = \int_A y \varphi(\mathbf{x}, y) d\lambda(y) = \psi(h(\mathbf{x})) \qquad (2)$$

for all $\mathbf{x}$ in $\Omega$. where $\psi(\cdot) \equiv \int_A y \pi(\cdot, y) d\lambda(y) : \Re \mapsto \psi(\Re)$ is assumed to be one-one and continuously differentiable. The inverse of $\psi(\cdot)$ is called the link function. In addition, define the second moment link function $\upsilon(\cdot) : \Re \mapsto \Re$ by $\upsilon(\cdot) \equiv \int_A y^2 \pi(\cdot, y) d\lambda(y)$. Assume that $\upsilon(\cdot)$ is continuous.

Denote the set of all such functions $\varphi(\cdot, \cdot) = \pi(h(\cdot), \cdot)$ by $\Phi$. This is the set of target functions that we will consider to approximate.

All density functions $\varphi$ in $\Phi$ have conditional mean function $\mu$'s belonging to $\psi(W_2^\infty)$, a transformed version of the Sobolev space $W_2^\infty$.

### 2.2 The Family of HME of GLM1s

An approximator in the HME family is assumed to have the following form:

$$f_\Lambda(\mathbf{x}, y; \theta) = \sum_{J \in \Lambda} g_J(\mathbf{x}; \mathbf{v}) \pi(\alpha_J + \beta_J^T \mathbf{x}, y), \qquad (3)$$

where $\pi(\cdot, \cdot)$ is as defined in Section 2.1. The parameters of this model include $\alpha_J \in \Theta_\alpha \subset \Re$ and $\beta_J \in \Theta_\beta \subset \Re^s$ with $\Theta_\alpha$ and $\Theta_\beta$ being some compact sets, as well as $\mathbf{v}$ which is some parameter for the gating function $g_J$'s. We use the symbol $\theta$ to represent the grand vector of parameters containing all the components of the parameters $\mathbf{v}$, $\alpha_J$ and $\beta_J$ for all $J \in \Lambda$. In (3), $\Lambda$ is the set of labels of all the experts in a network, referred to as a *structure*. Two quantities are associated with a structure: the dimension $\ell = \dim(\Lambda)$, which is the number of layers; and the cardinality $m = \text{card}(\Lambda)$, which is the number of experts. An HME of $\ell$-layers has a structure of the form $\Lambda = \otimes_{k=1}^\ell A_k$ where $A_k \subset \mathcal{N}$, $k = 1, \ldots, \ell$. (We use $\mathcal{N}$ to denote the set of all positive integers.) Note that in this paper we restrict attention to "rectangular-shaped" structures. A generic expert label $J$ in $\Lambda$ can then be expressed as $J = (j_1, \ldots, j_\ell)$ where $j_k \in A_k$ for each $k$.

To characterize a structure $\Lambda$, we often claim that it belongs to a certain *set of structures*. We now introduce three such sets of structures, $\mathcal{J}$, $\mathcal{J}_m$ and $\mathcal{S}$, which will be used later when formulating the results. The set of all possible HME structures under consideration is $\mathcal{J} = \{\Lambda : \Lambda = \otimes_{k=1}^{\ell} A_k; A_1, \ldots, A_\ell \subset \mathcal{N}; \ell \in \mathcal{N}\}$. The set of all HME structures containing no more than $m$ experts is denoted as $\mathcal{J}_m = \{\Lambda : \Lambda \in \mathcal{J}, \text{card}(\Lambda) \leq m\}$. We also introduce a symbol $\mathcal{S}$ to denote a generic subset of $\mathcal{J}$. This is introduced in order to formulate a major condition for the results of this paper to hold. This condition, to be formulated in the next section, will be specific to a generic subset $\mathcal{S}$ of HME structures.

Associated with a structure $\Lambda$ is a family of vectors of gating functions. Each member is called a *gating vector* and is labeled by a parameter vector $\mathbf{v} \in V_\Lambda$, $V_\Lambda$ being some parameter space specific to the structure $\Lambda$. Denote a generic gating vector as $G_{\mathbf{v},\Lambda} \equiv (g_J(\cdot; \mathbf{v}))_{J \in \Lambda}$. We assume the $g_J(\mathbf{x}; \mathbf{v})$'s to be nonnegative, with sum equal to unity, and continuous in $\mathbf{x}$ and $\mathbf{v}$. Note that $\int_A f_\Lambda(\mathbf{x}, y; \theta) d\lambda(y) = 1$ is ensured. Let $\mathcal{G} = \{G_{\mathbf{v},\Lambda} : \mathbf{v} \in V_\Lambda, \Lambda \in \mathcal{J}\}$ be the family of gating vectors, which will be referred to as the *gating class*.

Now we are ready to define the family of approximator functions. Let $\Pi_\Lambda$ be the set of all function $f_\Lambda$'s of the form (3), specific to a structure $\Lambda$, which can be denoted as $\Pi_\Lambda = \{f_\Lambda(\cdot, \cdot; \theta) : \theta \in \tilde{\Theta}_\Lambda\}$ where we use $\tilde{\Theta}_\Lambda$ to denote the set of all the grand parameter vector $\theta$'s. This set $\Pi_\Lambda$ is the set of HME functions from which an optimal function is chosen by the maximum likelihood method to approximate the truth. It is assumed that a structure $\Lambda$ is chosen *a priori*. In practice, people often analyze data using different choices of structures and select the best fitting model. We consider in this paper choosing among the set of structures $\mathcal{J}_m \cap \mathcal{S}$. Denote

$$\Pi_{m,\mathcal{S}} = \{f : f \in \Pi_\Lambda; \Lambda \in \mathcal{J}_m \cap \mathcal{S}\}. \quad (4)$$

This set, $\Pi_{m,\mathcal{S}}$, is the family of HME functions for which we examine the approximation rate in $\Phi$, as $m \to \infty$. Note that this family of HME functions is specific to $m$, the maximum number of experts, as well as to some subset $\mathcal{S}$ of HME structures, which will be specified later. We do not explicitly require that $\Pi_{m,\mathcal{S}}$ be a subset of $\Phi$ in this paper.

Each HME density function $f_\Lambda(\mathbf{x}, \cdot; \theta)$ generates a mean function $\mu_\Lambda(\mathbf{x}; \theta)$ by

$$\begin{aligned} \mu_\Lambda(\mathbf{x}; \theta) &= \int_A y f_\Lambda(\mathbf{x}, \cdot; \theta) d\lambda(y) \\ &= \sum_{J \in \Lambda} g_J(\mathbf{x}; \mathbf{v}) \psi(\alpha_J + \mathbf{x}^T \beta_J), \quad (5) \end{aligned}$$

where $\psi(\cdot) = \int_A y\pi(\cdot, y) d\lambda(y)$.

The parameterization of the HME functions is not identifiable, in the sense that two different parameters $\theta$ in $\tilde{\Theta}_\Lambda$ can represent the same density function $f$ in $\Pi_{m,\mathcal{S}}$. First, the density functions are invariant under permutation of the expert label $J$'s. Second, if two experts $J$ and $J'$ propose the same output, i.e., if $\alpha_J = \alpha_{J'}$ and $\beta_J = \beta_{J'}$, then the mixing proportions for these two experts can be arbitrary, as long as the sum of the two weights are unchanged. This can lead to the non-identifiability of some components of parameter $\mathbf{v}$. Our description of the estimation procedure and the statement of the results will take these identifiability issues into account. The identifiability issues also suggest that it makes more sense to formulate the consistency problem in terms of the predicted mean response, rather than to look at the consistency of the parameter estimates.

### 2.3 Method of Fitting

We will use the maximum likelihood method. Suppose we estimate the mean response $\mu(\mathbf{x})$ based on a data set of $n$ predictor-response pairs $(\mathbf{X}_i, Y_i)$, $\mathbf{X}_i \in \Omega$, $Y_i \in A$, $i = 1, \ldots, n$. Let the measure spaces $(\Omega, \mathcal{F}_\Omega, \kappa)$ and $(A, \mathcal{F}_A, \lambda)$ be as introduced in Section 2.1. Assume that $(\mathbf{X}_i, Y_i)$, $i = 1, \ldots, n$ are independent and identically distributed (i.i.d.) random vectors. The probability measure for $\mathbf{X}_i$ is $\kappa$. The probability measure of $Y_i$ conditional on $\mathbf{X}_i = \mathbf{x}$ has a density $\varphi(\mathbf{x}, \cdot)$ with respect to the measure $\lambda$, for all $\mathbf{x} \in \Omega$.

The log-likelihood function based on the HME model is

$$\mathrm{L}_{n,\Lambda}(\theta; \omega) = n^{-1} \sum_{i=1}^n \log\{f_\Lambda(\mathbf{X}_i, Y_i; \theta) / \varphi_0(\mathbf{X}_i, Y_i)\} \quad (6)$$

where $f_\Lambda(\cdot, \cdot; \theta) \in \Pi_\Lambda$ is defined in Section 2.2, $\theta \in \tilde{\Theta}_\Lambda$, $\omega$ is the stochastic sequence of events $(\mathbf{X}_i, Y_i)$, $i = 1, \ldots$, and $\varphi_0(\mathbf{X}_i, Y_i)$ can be any positive measurable function of the observed data that does not depend on the parameter $\theta$. Define the maximum likelihood estimator (MLE) $\hat{\theta}_{n,\Lambda}(\omega)$ to be a maximizer (can be one out of many) of $\mathrm{L}_{n,\Lambda}(\theta; \omega)$ over a compact set $\bar{B}_\Lambda \subset \tilde{\Theta}_\Lambda$, i.e.,

$$\hat{\theta}_{n,\Lambda}(\omega) = \arg\max_{\theta \in \bar{B}_\Lambda} \{\mathrm{L}_{n,\Lambda}(\theta; \omega)\}. \quad (7)$$

The maximum likelihood method, in the large sample size limit, essentially searches for $\theta$ which minimizes the KL divergence $\mathrm{KL}(f_\Lambda, \varphi)$ between $f_\Lambda =$

$f_\Lambda(\cdot,\cdot;\theta) \in \Pi_\Lambda$ and $\varphi = \varphi(\cdot,\cdot) \in \Phi$, where

$$\mathrm{KL}(f,g) \equiv \int_{\Omega \otimes A} g(\mathbf{x},y) \log\left\{\frac{g(\mathbf{x},y)}{f(\mathbf{x},y)}\right\} d\kappa(\mathbf{x})d\lambda(y). \quad (8)$$

Due to the non-identifiability of the parameterization, there is *a set* of $\theta$'s in $\bar{B}_\Lambda$ that minimize the KL divergence. Denote this set as $\Theta_\Lambda$, which could be expressed as

$$\Theta_\Lambda = \{\theta \in \bar{B}_\Lambda : \theta = \arg\min_{\theta^* \in \bar{B}_\Lambda} \mathrm{KL}(f_\Lambda(\cdot,\cdot;\theta^*), \varphi)\}. \quad (9)$$

Based on *any* MLE $\hat{\theta}_{n,\Lambda} = \hat{\theta}_{n,\Lambda}(\cdot)$, a predicted mean response can be constructed as $\mu_\Lambda(\mathbf{x}; \hat{\theta}_{n,\Lambda})$. We do not explicitly require that for two different global MLEs the predicted mean responses be the same. The MSE of *a* predicted mean response is defined by

$$(\mathrm{MSE})_{n,\Lambda} = \mathrm{E} \int \{\mu_\Lambda(\mathbf{x}; \hat{\theta}_{n,\Lambda}) - \mu(\mathbf{x})\}^2 d\kappa(\mathbf{x}), \quad (10)$$

where E is the expectation taken on the MLE $\hat{\theta}_{n,\Lambda}$, $\mu_\Lambda$ and $\mu$ are defined in (5) and (2), respectively.

### 2.4 Technical Definitions

Two technical definitions are introduced below. We will use these definitions to formulate a major condition under which our theorem holds.

**Definition 1** *(Fine Partition).* For $\nu = 1, 2, \ldots$, let $\mathbf{Q}^{(\nu)} = \{Q_J^{(\nu)}\}_{J \in \Lambda^{(\nu)}}$, $\Lambda^{(\nu)} \in \mathcal{J}$, be a partition of $\Omega \subset \Re^s$ with Euclidean metric $\rho(\cdot,\cdot)$. (This means that for fixed $\nu$, the $Q_J^{(\nu)}$'s are mutually disjoint subsets of $\Re^s$ whose union is $\Omega$.)

Let $p_\nu = \mathrm{card}(\Lambda^{(\nu)})$, $(p_\nu \in \mathcal{N})$.

If $p_\nu \to \infty$ and for all $\xi, \eta \in Q_J^{(\nu)}$, $\rho(\xi, \eta) \le c_0/p_\nu^{1/s}$ for some constant $c_0$ independent of $\nu$, $J$, $\xi$, $\eta$, then $\{\mathbf{Q}^{(\nu)} : \nu = 1, 2, \ldots\}$ is called a sequence of fine partitions with structure sequence $\{\Lambda^{(\nu)}\}$ and cardinality sequence $\{p_\nu\}$.

**Definition 2** *(Sub-Geometric).* A sequence $\{p_\nu\}$ is sub-geometric if $p_\nu \in \mathcal{N}$, $p_\nu \to \infty$ as $\nu \to \infty$, and $1 < |p_{\nu+1}/p_\nu| < M_2$ for all $\nu = 1, 2, \ldots$, where $M_2$ is some finite constant.

## 3 Results and Conditions

In this paper, we will only state the conditions and the results. The proofs of the results will appear elsewhere.

**Condition 1** *(Scope of Maximum Likelihood Searching)* The scope of the maximum likelihood searching, $\bar{B}_\Lambda$, is a compact set which is so large that it contains a point $\theta_\Lambda^{\mathrm{LS}}$ which minimizes the following $L_2$ distance between $f_\Lambda(\cdot,\cdot;\theta) \in \Pi_\Lambda$ and $\varphi(\cdot,\cdot) \in \Phi$ among all $\theta \in \tilde{\Theta}_\Lambda$:

$$\|f_\Lambda(\cdot,\cdot;\theta) - \varphi(\cdot,\cdot)\|_2 = \int_{\Omega \otimes A} \{f_\Lambda(\mathbf{x},y;\theta) - \varphi(\mathbf{x},y)\}^2 \varphi(\mathbf{x},y) d\kappa d\lambda.$$

I.e., $\bar{B}_\Lambda$ is chosen so large that it contains a point $\theta_\Lambda^{\mathrm{LS}}$ satisfying

$$\|f_\Lambda(\cdot,\cdot,\theta_\Lambda^{\mathrm{LS}}) - \varphi(\cdot,\cdot)\|_2 = \inf_{\theta \in \tilde{\Theta}_\Lambda} \|f_\Lambda(\cdot,\cdot,\theta) - \varphi(\cdot,\cdot)\|_2.$$

**Condition 2** *(Uniform Convergence of Log-Likelihood)* For some function $\varphi_0(\cdot,\cdot)$ introduced in (6),

$$\sup_{\theta \in \bar{B}_\Lambda} |L_{n,\Lambda}(\theta; \omega) - L_{\infty,\Lambda}(\theta)| \to 0$$

for almost all stochastic sequences $\omega$, where

$$L_{\infty,\Lambda}(\theta) = \int_{\Omega \otimes A} \varphi(\mathbf{x},y) \log\{f_\Lambda(\mathbf{x},y;\theta)/\varphi_0(\mathbf{x},y)\} d\kappa d\lambda$$

is continuous in $\theta \in \bar{B}_\Lambda$.

**Condition 3** *($A_{\mathcal{S},p}$).* For a subset $\mathcal{S} \subset \mathcal{J}$, there is a fine partition sequence $\{\{Q_J^{(\nu)}\}_{J \in \Lambda_0^{(\nu)}} : \nu = 1, 2, \ldots\}$ with a sub-geometric cardinality sequence $\{p_\nu : \nu = 1, 2, \ldots\}$ and a structure sequence $\{\Lambda_0^{(\nu)} : \nu = 1, 2, \ldots\}$ where $\Lambda_0^{(\nu)} \in \mathcal{S}$ for all $\nu$, such that for all $\nu$, for all $\varepsilon > 0$, there exists $\mathbf{v}_\varepsilon \in V_{\Lambda_0^{(\nu)}}$ and a gating vector

$$G_{\mathbf{v}_\varepsilon, \Lambda_0^{(\nu)}} = \{g_J(\mathbf{x}; \mathbf{v}_\varepsilon)\}_{J \in \Lambda_0^{(\nu)}} \in \mathcal{G}, \quad \Lambda_0^{(\nu)} \in \mathcal{S},$$

such that

$$\sup_{J \in \Lambda_0^{(\nu)}} \|g_J(\cdot; \mathbf{v}_\varepsilon) - \chi_{Q_J^{(\nu)}}(\cdot)\|_{p,\kappa} \le \varepsilon. \quad (11)$$

Here, $\|f(\cdot)\|_{p,\kappa} \equiv \{\int_\Omega |f(\mathbf{x})|^p d\kappa(\mathbf{x})\}^{1/p}$, where $\kappa$ is any finite measure on $\Omega$; $\chi_B(\cdot)$ is the character function for a subset $B$ of $\Omega$, i.e., $\chi_B(\mathbf{x}) = 1$ if $\mathbf{x} \in B$, 0 otherwise.

This condition is a restriction on the gating class $\mathcal{G}$. Loosely speaking, it indicates that the vectors of local gating functions in the parametric family should arbitrarily approximate the vector of characteristic functions for a partition of the predictor space $\Omega$, as the cells of the partition become finer.

The next two conditions place restrictions on the function $\pi(\cdot,\cdot)$ introduced in (1).

**Condition 4** *(One-Sided Boundedness of $\pi$ and $\partial_h \pi$)* *For any bounded subset $H$ of $\Re$, there is a constant $M_1 < \infty$, possibly depending on $H$, such that*

*(a)* $\sup_{h \in H, y \in A} |\pi(h, y)| \leq M_1$

*(b)* $\sup_{h \in H, y \in A} |\partial_h \pi(h, y)| \leq M_1$

**Condition 5** *(Two-Sided Boundedness of $\pi$)* *For any bounded subset $H$ of $\Re$, there exist constants $M_1, M_3 \in (0, \infty)$, possibly dependent on $H$, such that*

$$M_3 \leq |\pi(h, y)| \leq M_1 \text{ for all } h \in H, y \in A.$$

[Note that this is a stronger condition than Condition 4(a).]

Now we are ready to state our main results.

The following theorem states that the HME of GLM1s can be used to approximate one-parameter exponential family densities with arbitrary smooth mean functions in a transformed Sobolev space $\psi(W_2^\infty)$, as the number of experts $m$ increases.

**Theorem 1** *(Approximation Rate). Under the condition $A_{\mathcal{S},p}$ and Condition 4,*

$$\sup_{\varphi \in \Phi} \inf_{f \in \Pi_{m,\mathcal{S}}} \|f - \varphi\|_p \leq \frac{c}{m^{2/s}}$$

*for some constant $c > 0$. Here $\|f\|_p \equiv \left\{ \int_{\Omega \otimes A} |f(\mathbf{x}, y)|^p d\sigma(\mathbf{x}, y) \right\}^{1/p}$ where $\sigma$ is any probability measure on $\Omega \otimes A$, such that $f, \varphi$ are measurable for all $f \in \cup_{\Lambda \in \mathcal{J}} \Pi_\Lambda$ and $\varphi \in \Phi$, and $\sigma$ has a density function $\tilde{\varphi}(\mathbf{x}, y)$ with respect to the product measure $\kappa \otimes \lambda$, such that $\int_A \tilde{\varphi}(\mathbf{x}, y) d\lambda(y) = 1$ for all $\mathbf{x} \in \Omega$.*

It is common to measure the discrepancy between two density functions by the KL divergence. The following theorem states that the HME functions can approach the target functions in the sense that the KL divergence converges to zero, as the number of experts $m$ increases.

**Theorem 2** *(Approximation Rate in KL Divergence). Under Conditions 4, 5 and the condition $A_{\mathcal{S},2}$,*

$$\sup_{\varphi \in \Phi} \inf_{f \in \Pi_{m,\mathcal{S}}} \text{KL}(f, \varphi) \leq c^*/m^{4/s}$$

*for some constant $c^* > 0$, where $\text{KL}(\cdot, \cdot)$ is defined in (8).*

The next theorem states that the maximum likelihood method based on GLM1 models is consistent in estimating the mean functions in $\psi(W_2^\infty)$.

**Theorem 3** *(Consistency of the Maximum Likelihood Method) Let $(\text{MSE})_{n,\Lambda}$ be as defined in (10). Under regularity conditions 1, 2, 4, 5 and $A_{\mathcal{S},2}$,*

$$\lim_{m \to \infty} \limsup_{n \to \infty} \inf_{\Lambda \in \mathcal{S} \cap \mathcal{J}_m} (\text{MSE})_{n,\Lambda} = 0.$$

*Here $s = \dim(\Omega)$, $n$ is the sample size, $m = \sup_{\Lambda \in \mathcal{S} \cap \mathcal{J}_m} \{\text{card}(\Lambda)\}$, and $\mathcal{J}_m = \{\Lambda : \Lambda \in \mathcal{J}, \text{card}(\Lambda) \leq m\}$ is the set of all HME structures containing no more than $m$ experts. Actually*

$$\limsup_{n \to \infty} \inf_{\Lambda \in \mathcal{S} \cap \mathcal{J}_m} (\text{MSE})_{n,\Lambda} \leq \frac{\bar{c}}{m^{4/s}},$$

*where $\bar{c}$ is a positive constant independent of $n$, $m$ and the structure $\Lambda$.*

Next we claim that the commonly used logistic type gating vectors [e.g., in Jordan and Jacobs (1994)] satisfy the condition $A_{\mathcal{S},p}$ for some $\mathcal{S}$ and $p$. We first define the logistic gating class $\mathcal{L}$.

**Definition 3** *(Logistic Gating Class). For $J = (j_1, \ldots, j_\ell) \in \Lambda$, $\Lambda = \otimes_{k=1}^\ell A_k \in \mathcal{J}$, Let*

$$g_J(\mathbf{x}; \mathbf{v}) = g_{j_1 \ldots j_\ell}(\mathbf{x}; \mathbf{v})$$
$$= \prod_{q=1}^\ell \frac{\exp(\phi_{j_1 \ldots j_{q-1} j_q} + \mathbf{x}^T \boldsymbol{\gamma}_{j_1 \ldots j_{q-1} j_q})}{\sum_{k_q \in A_q} \exp(\phi_{j_1 \ldots j_{q-1} k_q} + \mathbf{x}^T \boldsymbol{\gamma}_{j_1 \ldots j_{q-1} k_q})},$$

$\mathbf{x} \in \Omega = [0, 1]^s$, $\boldsymbol{\gamma}_{j_1 \ldots j_{q-1} j_q} \in \Re^s$, $\phi_{j_1 \ldots j_{q-1} j_q} \in \Re$, *for all $j_r \in A_r$; $r = 1, \ldots, q$; $q = 1, \ldots, \ell$;*

$$\mathbf{v} = \{(\boldsymbol{\gamma}_{j_1 \ldots j_{q-1} j_q}, \phi_{j_1 \ldots j_{q-1} j_q}) : j_r \in A_r; r = 1, \ldots, q; q = 1, \ldots, \ell\}.$$

*Let $V_\Lambda$ be the set of all such $\mathbf{v}$'s. Then, $G_{\mathbf{v},\Lambda} \equiv \{g_J(\cdot; \mathbf{v})\}_{J \in \Lambda}$ is called a vector of logistic gating functions for structure $\Lambda$. The set of all such $G_{\mathbf{v},\Lambda}$'s, $\mathbf{v} \in V_\Lambda$, $\Lambda \in \mathcal{J}$, is denoted as $\mathcal{L}$, the logistic gating class.*

For the logistic gating class, we have the following lemma.

**Lemma 1** *For HME with logistic gating class $\mathcal{G} = \mathcal{L}$, the condition $A_{\mathcal{S},p}$ is satisfied for all $p \in \mathcal{N}$, for all finite measure $\kappa$ associated with the $L_p$-norm, which is absolutely continuous with respect to the Lebesgue measure on $\Omega$, and for $\mathcal{S} = \mathcal{S}_s$ where $\mathcal{S}_s \equiv \{\Lambda \in \mathcal{J} : \dim(\Lambda) \leq s\}$, $s = \dim(\Omega)$.*

From this lemma, we immediately obtain the following corollary:

**Corollary 1** *If the gating class is $\mathcal{G} = \mathcal{L}$ (logistic), then Theorem 1, 2 and 3 hold for $\mathcal{S} = \mathcal{S}_s$ and any $p \in \mathcal{N}$ where $\mathcal{S}_s = \{\Lambda \in \mathcal{J} : \dim(\Lambda) \leq s\}$, $s = \dim(\Omega)$.*

This corollary indicates that the approximation rates and the consistency result can be obtained within the family of HME networks with no more than $s$ layers, $s$ being the dimension of the predictor.

We conclude this section by making some remarks on the conditions 1, 2, $A_{\mathcal{S},p}$, 4 and 5.

(i) Only Conditions 2 and $A_{\mathcal{S},p}$ are "big" conditions. $A_{\mathcal{S},p}$ was checked for logistic gating class. Condition 2 could be reduced to more primitive conditions by a uniform law of large numbers (Jennrich 1969, Theorem 2). It is straightforward to show that Condition 2 will hold if the following more primitive condition holds:

**Condition 6** *(Alternative of Condition 2).* *For any bounded subset $H \subset \Re$, there exists some integrable function $M(y)$ (i.e., $E\{M(Y)\} < \infty$,) such that*

$$\sup_{h \in H} |\log\{\pi(h, y)/\pi_0(y)\}| \leq M(y)$$

*for some positive measurable function $\pi_0(y)$ independent of $h$.*

This condition is satisfied for one-parameter exponential family regression models, in which case Condition 6 is reduced to the existence of $E(Y)$, the unconditional expectation of the response variable.

(ii) Condition 5, used in Theorems 2 and 3, holds for logistic regression models, but does not hold for normal, exponential or Poisson models, where the density functions can be arbitrarily close to zero in the tails of the distribution of $y$. However the condition does hold for truncated normal, exponential and Poisson models, and the truncation parameters can be made arbitrarily large. In some sense, the untruncated exponential family regression models might be regarded as being very "close" to the truncated ones with very large truncation parameters, for which Condition 5 holds.

(iii) The other conditions are very mild. For instance, Condition 4 is easily checked to hold for normal, exponential and Poisson distributions. Nevertheless, Condition 1 is hard to check in practice, although it looks plausible.

## 4 Conclusions

We investigated the power of the HME networks of one parameter exponential family regression models with generalized linear mean functions (GLM1 experts) in terms of approximating a certain class of relatively arbitrary density functions, namely, the density functions of one-parameter exponential family regression models with conditional mean functions belonging to a transformed Sobolev space. We demonstrated that the approximation rate is of order $O(m^{-2/s})$ in $L_p$ norm, or of order $O(m^{-4/s})$ in KL divergence. We also showed that the maximum likelihood (ML) approach, which is associated with some optimal statistical properties and a convenient maximization algorithm, is consistent in estimating the mean response from data, as the sample size and the number of experts both increase. Moreover, we claim that the approximation rate and the consistency result can be achieved within the family of HME structures with no more than $s$ layers, where $s$ is the dimension of the predictor, and $m$ is the number of experts in the network. Two remarks can be made: (i) We do not claim that the $O(m^{-2/s})$ approximation rate in $L_p$ norm cannot be achieved by fewer than $s$ layers of experts. (ii) We do not claim that the $O(m^{-2/s})$ rate is optimal. In fact, for the special case of mixing linear model experts, Zeevi et al. (1998) have shown that the rate $O(m^{-2/s})$ can be achieved within one-layer networks. Also, they have shown that a better rate can be achieved if higher-than second-order continuous differentiability of the target functions is assumed. Our work are different from Zeevi et al. (1998) on the following aspects: (i) We deal with mixtures of *generalized* linear models instead of the mixtures of ordinary linear models. (ii) We consider the set-up of the HME networks instead of the single-layer mixtures of experts. (iii) We consider the maximum likelihood method instead of the least-squares approach for model fitting. (iv) Related to the use of the maximum likelihood method, we obtained the approximation rate in terms of probability density functions instead of in terms of the mean response. (v) We have formulated the conditions of our results in a way that is protective of the inherent non-identifiability problems of the parameterization.

## Acknowledgments

The authors wish to thank the referees for helpful comments in improving the presentation of this paper. Martin A. Tanner was supported in part by NIH Grant CA35464.

## References

BICKEL, P. J., AND DOKSUM, K. A. (1977). *Mathematical Statistics*, Prentice-Hall, Englewood Cliffs, New Jersey.

BISHOP, C .M. (1995). *Neural Networks for Pattern Recognition.* Oxford University Press, New York.

CACCIATORE, T. W. AND NOWLAN, S. J. (1994). Mixtures of controllers for jump linear and nonlinear plants. In G. Tesauro, D.S. Touretzky, and T.K. Leen (Eds.), *Advances in Neural Informations Processing Systems 6.* Morgan Kaufmann, San Mateo, CA.

GHAHRAMANI, Z. AND HINTON, G. E. (1996). The EM algorithm for mixtures of factor analyzers. Technical Report CRG-TR-96-1, Department of Computer Science, University of Toronto, Toronto, Ontario.

HAYKIN, S. (1994). *Neural Networks.* Macmillan College Publishing Company, New York.

JAAKKOLA, T. S. AND JORDAN, M. I. (1998). Improving the mean field approximation via the use of mixture distributions. In M.I. Jordan (Ed.), *Learning in Graphical Models.* Kluwer Academic Publishers.

JACOBS, R. A., JORDAN, M. I., NOWLAN, S. J., AND HINTON, G. E. (1991). Adaptive mixtures of local experts. *Neural Comp.* **3** 79-87.

JENNRICH, R. I. (1969). Asymptotic properties of nonlinear least squares estimators. *Ann. of Math. Stat.* **40** 633-643.

JORDAN, M. I., AND JACOBS, R. A. (1994). Hierarchical mixtures of experts and the EM algorithm. *Neural Comp.* **6** 181-214.

JORDAN, M. I., AND XU, L. (1995). Convergence results for the EM approach to mixtures-of-experts architectures. *Neural Networks* **8** 1409-1431.

MCCULLAGH, P., AND NELDER, J. A. (1989). *Generalized Linear Models*, Chapman and Hall, London.

MEILĂ, M. AND JORDAN, M. I. (1995). Learning fine motion by Markov mixtures of experts. A.I. Memo No. 1567, Artificial Intelligence Laboratory, Massachusetts Institute of Technology, Cambridge, MA.

MHASKAR, H. N. (1996). Neural networks for optimal approximation of smooth and analytic functions. *Neural Comp.* **8** 164-177.

PENG, F., JACOBS, R. A., AND TANNER, M. A. (1996). Bayesian inference in mixtures-of-experts and hierarchical mixtures-of-experts models with an application to speech recognition. *J. Amer. Statist. Assoc.* **91** 953-960.

TIPPING, M. E. AND BISHOP, C. M. (1997). Mixtures of probabilistic principal component analysers. Technical Report NCRG-97-003, Department of Computer Science and Applied Mathematics, Aston University, Birmingham, UK.

TITTERINGTON, D. M., SMITH, A. F. M., AND MAKOV, U. E. (1985). *Statistical Analysis of Finite Mixture Distributions.* John Wiley, New York.

WHITE, H. (1994). *Estimation, Inference and Specification Analysis*, Cambridge University Press, Cambridge, England.

ZEEVI, A., MEIR, R., AND MAIOROV, V. (1998). Error bounds for functional approximation and estimation using mixtures of experts. *IEEE Trans. Information Theory* (To appear.)